\documentclass[conference]{IEEEtran}
\IEEEoverridecommandlockouts
\usepackage{cite}
\usepackage{amsmath,amssymb,amsfonts}
\usepackage{algorithmic}
\usepackage{graphicx}
\usepackage{textcomp}
\usepackage{xcolor}
\usepackage{bm}% bold math
\def\BibTeX{{\rm B\kern-.05em{\sc i\kern-.025em b}\kern-.08em
    T\kern-.1667em\lower.7ex\hbox{E}\kern-.125emX}}
\begin{document}

\title{Controlling dynamical systems to complex target states using machine learning: next-generation vs. classical reservoir computing\\

}

\author{\IEEEauthorblockN{1\textsuperscript{st} Alexander Haluszczynski}
\IEEEauthorblockA{\textit{risklab} \\
\textit{Allianz Global Investors}\\
Munich, Germany \\
alexander.haluszczynski@gmail.com}
\and
\IEEEauthorblockN{2\textsuperscript{nd} Daniel Köglmayr}
\IEEEauthorblockA{\textit{Institut f{\"u}r KI Sicherheit} \\
\textit{Deutsches Zentrum f{\"u}r Luft- und Raumfahrt (DLR)}\\
Ulm, Germany \\
daniel.koeglmayr@dlr.de}
\and

\IEEEauthorblockN{3\textsuperscript{rd} Christoph R{\"a}th}
\IEEEauthorblockA{\textit{Institut f{\"u}r KI Sicherheit} \\
\textit{Deutsches Zentrum f{\"u}r Luft- und Raumfahrt (DLR)}\\
Ulm, Germany \\
christoph.raeth@dlr.de}
}

\maketitle

\begin{abstract}
Controlling nonlinear dynamical systems using machine learning allows to not only drive systems into simple behavior like periodicity but also to more complex arbitrary dynamics. For this, it is crucial that a machine learning system can be trained to reproduce the target dynamics sufficiently well. On the example of forcing a chaotic parametrization of the Lorenz system into intermittent dynamics, we show first that classical reservoir computing excels at this task. In a next step, we compare those results based on different amounts of training data to an alternative setup, where next-generation reservoir computing is used instead. It turns out that while delivering comparable performance for usual amounts of training data, next-generation RC significantly outperforms in situations where only very limited data is available. This opens even further practical control applications in real world problems where data is restricted.
\end{abstract}


\section{Introduction}
It has been a significant breakthrough to be able to control chaotic systems into stable dynamical states \cite{ott1990controlling,shinbrot1993using}. Classical approaches include delayed feedback control \cite{pyragas1992continuous} as well as OGY control \cite{ott1990controlling} where a system in a chaotic state is driven into a fixed point or periodic orbit by applying an external force. While most approaches were only able to control systems into rather simple dynamical states, it has been shown recently that more complex target states can be achieved by leveraging a machine learning based control mechanism \cite{haluszczynski2021controlling} without requiring knowledge of the underlying equations. In order to derive the control force, a good prediction of the desired dynamics is needed. 

There has been remarkable progress made in the prediction of chaotic systems in the recent past with reservoir computing standing (RC) out \cite{maass2002real,jaeger2004harnessing} delivering high accuracy predictions while being less data hungry than comparable methods due to its efficient architecture. There have been many advances and refinements of RC (e.g. see \cite{carroll2022time}, \cite{gallicchio2018design}, \cite{kong2021machine}) mounting to the recent proposal of next-generation reservoir computing (NG-RC)  \cite{gauthier2021next}, where the randomly generated underlying neural network function is replaced by a determined polynomial multiplication rule of time-shifted inputs.

In this study we compare the results of the reservoir computing powered control mechanism \cite{haluszczynski2021controlling} to a similar setup, where next-generation reservoir computing is used instead. it is shown that the required amount of training data can be significantly lowered roughly by a factor of 10. Requiring only very few training data, this approach becomes feasible for even more potential practical applications.

\section{Controlling Chaos}\label{sec:controlIntro}

The original idea of chaos control is that unstable periodic orbits can be stabilized by small perturbations of the system resulting from applying an external force. Traditional approaches like OGY control \cite{ott1990controlling} and delayed feedback control \cite{pyragas1992continuous} either require knowledge of the underlying equations of the system or large amounts of data due to relying on phase space methods. Furthermore, it is only possible to control the system in simple dynamical target states. While there are many extensions \cite{boccaletti2000control} including a way to "chaotify" periodic or synchronized dynamics \cite{schiff1994controlling}, the ability to achieve arbitrary dynamical target states only emerged recently \cite{haluszczynski2021controlling}. 

The idea is to train a machine learning system -- reservoir computing in this case -- on the original desired dynamics of the system characterized by state $\textbf{X}$. If then the behavior of the system changes to a new state $\textbf{Y}$ resulting from external factors or order parameter change, a force is applied such that the dynamics of the system are pushed back into the original state $\textbf{X}$. However, the causes for the change in behavior are still present such that the system would stay in its new state $\textbf{Y}$ in case no force is applied. Deriving a suitable control force $\textbf{F}(t)$ requires the knowledge of how the trajectory $\textbf{u}(t)$ of the system would have evolved if the system would still exhibit its original dynamics of state $\textbf{X}$ -- which we call $\textbf{v}(t)$. Therefore, if the reservoir computer was able to properly learn the dynamics of the system while being in its original state, one can predict how the trajectory would evolve in future, if the system was still in state $\textbf{X}$ instead of state $\textbf{Y}$. This can then be compared to the actual evolution of the system while being in state $\textbf{Y}$ such that the force results from the difference of the actual and hypothetical trajectory
\begin{eqnarray}
\textbf{F}(t)=K(\textbf{u}(t) - \textbf{v}(t)) \ ,
\label{eq:force}
\end{eqnarray} 
where $K$ is a parameter scaling the force. As the control force only depends on the measured actual coordinates as well as the predicted coordinates, no knowledge of the underlying equations of the system is needed and therefore the mechanism is applicable for many real world applications where a precise mathematical model is missing. However, for proper analysis we resort to a mathematical example system, which is simulated using its known equations. To apply the control force in this context, one has to simply add it to the differential equations $\dot{f}$ of the system. For solving the differential equations for the next time step one has to evaluate then 
\begin{eqnarray}
\textbf{u}(t+\Delta{t}) = \int_{t}^{t+\Delta{t}} (\dot{f}(\textbf{u}(\tilde{t})) + \textbf{F}(\tilde{t})) d\tilde{t} \ .
\end{eqnarray} 
We demonstrate the control mechanism for the Lorenz system \cite{lorenz1963deterministic}, which can exhibit periodic, intermittent or chaotic behavior depending on the parameter choice. The equations $\dot{f}(\textbf{u}(t)$ read 
\begin{eqnarray}
\dot x = \sigma (y-x); \ \ \dot y = x (\rho-z)-y;\ \ \dot z = x y - \beta z \ ,
\label{eq:lorenz}
\end{eqnarray} 
and $\bm{\pi} \equiv (\sigma, \rho, \beta)$ are the order parameters of the Lorenz system. The equations are solved using the 4th order Runge-Kutta method with a time resolution $\Delta t = 0.05$ and the scaling parameter for the control force is set to $K=1/\Delta{t}$.

\section{Predictions and measures}
\subsection{Classical Reservoir Computing}\label{classicRC}

A reservoir computer \cite{jaeger2001echo, maass02, jaeger2004harnessing} is a dynamical system with a recurrent neural network architecture. It consists of a typically randomly created internal network called reservoir $\textbf{A}$, which does not change over time and is characterized by its spectral radius $\rho$. While there are many possibilities to come up with such a network, a classical approach is to use sparse Erd{\"o}s-Renyi random networks \cite{erdos1959random}. Throughout this study we choose our network to have $D_{r}=300$ nodes that are connected with a probability $p=0.02$. While the network itself does not dynamically change, we define a state vector $\textbf{r}(t)$ that changes over time according to the recurrent equation
\begin{eqnarray}
\textbf{r}(t+ \Delta{t}) = tanh(\textbf{A}\textbf{r}(t) + \textbf{W}_{in} \textbf{u}(t)) \ .
\label{eq:updating}
\end{eqnarray} 
Here, $\textbf{W}_{in}$ is a $D_{r} \times D$ dimensional input mapping matrix, which is generated using uniformly distributed random numbers within a certain range [-$\omega$, $\omega$]. The job of $\textbf{W}_{in}$ is to feed the $D=3$ dimensional input data $\textbf{u}(t)$ of the Lorenz system into the $D_{r}=300$ dimensional state vector. In order to introduce nonlinearity, the hyperbolic tangent is used as the activation function. Output $\textbf{v}(t + \Delta{t})$ is created based on $\textbf{r}(t)$ by applying a linear readout matrix $\textbf{W}_{out}$ such that 
\begin{eqnarray}
\textbf{v}(t) = \textbf{W}_{out}(\mathbf{\tilde{r}}(t),  \textbf{P}) = \textbf{P}\mathbf{\tilde{r}}(t) \ , 
\label{eq:output}
\end{eqnarray}
where $\mathbf{\tilde{r}} = \{\mathbf{r}, \mathbf{r}^{2}\}$. The squared elements of the reservoir states are included to avoid problems caused by the anti-symmetry of the hyperbolic tangent as explained in \cite{herteux2020reservoir}. Determining the matrix $\textbf{P}$ is called training. This is achieved by recording a sufficient number of reservoir states $\textbf{r}(t_{w}...t_{w}+t_{T})$ and then choosing $\textbf{P}$ such that the difference between output $\textbf{v}$ of the reservoir and the known real data $\textbf{v}_{R}(t_{w}...t_{w}+t_{T})$ is minimized. As this procedure only involves the linear readout layer, training a reservoir computer is much faster as compared to other recurrent neural network based methods. Typically, an initial synchronization or washout phase is allowed meaning that a number of states $t_{w}$ are discarded for the training process. The optimization is done using Ridge regression and minimizes 
 \begin{eqnarray}
\sum^{}_{-T \leq t \leq 0} {\parallel  \textbf{W}_{out}(\mathbf{\tilde{r}}(t), \textbf{P}) - \textbf{v}_{R}(t) \parallel}^2 - \beta {\parallel \textbf{P} \parallel}^2 \ ,
\label{eq:minimizing}
\end{eqnarray} 
where $\beta$ is the regularization constant that punishes large values of the fitting parameters and thus avoids overfitting. Plugging the readout $\textbf{P}\mathbf{\tilde{r}}(t)$ into the recursive equation for the state vector $\textbf{r}(t)$ one can produce predictions of any length
\begin{eqnarray}
\begin{aligned}
\textbf{r}(t+ \Delta{t}) &= tanh(\textbf{A}\textbf{r}(t) + \textbf{W}_{in} \textbf{W}_{out}(\mathbf{\tilde{r}}(t),\textbf{P})) \\
                                  &= tanh(\textbf{A}\textbf{r}(t) + \textbf{W}_{in} \textbf{P}\mathbf{\tilde{r}}(t))  \ .
\label{eq:updatingprediction}
\end{aligned}
\end{eqnarray} 
In the result section, predictions are evaluated based on 10000 prediction time steps while we analyze the results for different amounts of training data and corresponding washout steps. In this study, we did not optimize for parameters ($\omega$, $\rho$, $\beta$), but heuristically set reasonable values based on previous research and insights from \cite{ haluszczynski2020reducing}, which are mentioned in the respective results sections.

\subsection{Next-Generation Reservoir Computing}\label{ngRC}
\begin{figure*}[t!] 
  \begin{center}
    \includegraphics[width=1.0\linewidth]{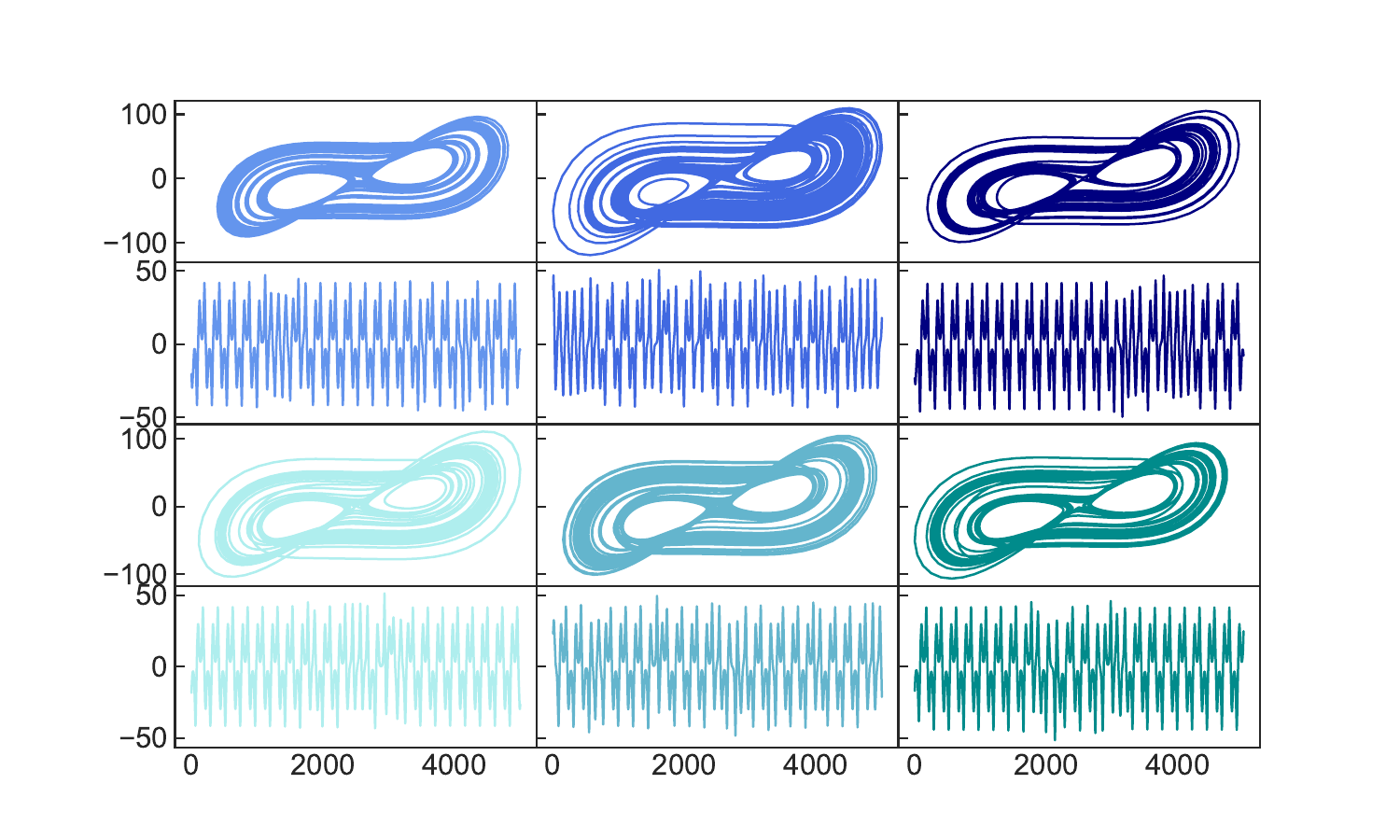}
    \caption{Chaotic to intermittent control. Top row and third row: 2D attractor representation in the x-y plane. Second row and bottom: X coordinate time series. The left plots show the original intermittent state, which changes to a chaotic state (middle) after tuning the order parameter. After applying the control mechanism, the system is forced into an intermittent state again (right). The first two rows reflect the results for classical RC, while the last two rows show results for NG-RC. In both cases, 5000 training steps were used.} 
    \label{fig:chaos2intermittent}
  \end{center}
\end{figure*}
\begin{figure*}[t!] 
  \begin{center}
    \includegraphics[width=1.0\linewidth]{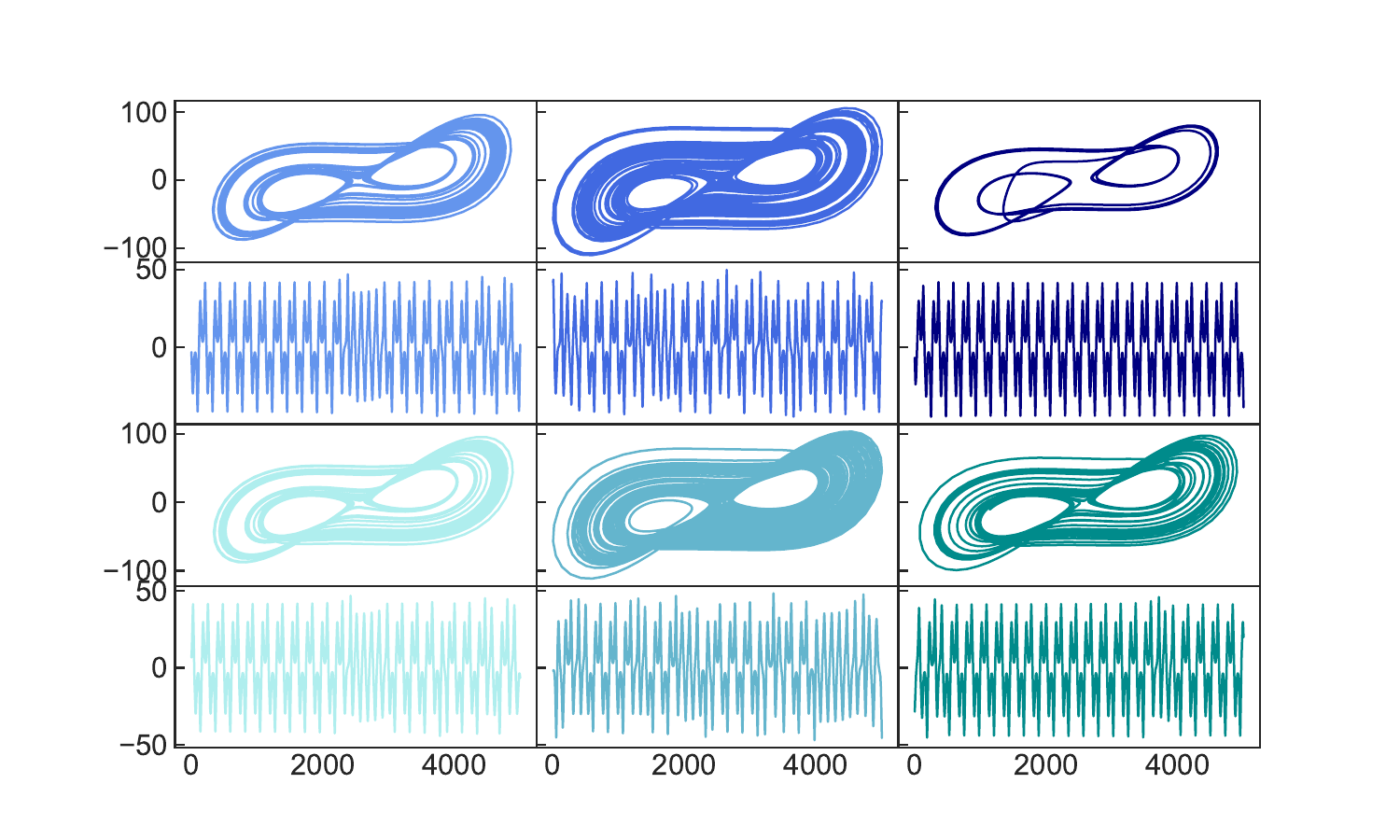}
    \caption{Chaotic to intermittent control. Top row and third row: 2D attractor representation in the x-y plane. Second row and bottom: X coordinate time series. Left plots show the original intermittent state which changes to a chaotic state (middle) after tuning the order parameter. After applying the control mechanism, the system is forced into an intermittent state again (right). The first two rows reflect the results for classical RC, while the last two rows show results for NG-RC. This time, only 500 training steps were used.} 
    \label{fig:chaos2intermittent500}
  \end{center}
\end{figure*}

Next-generation reservoir computing (NG-RC) \cite{gauthier2021next} is characterized by the absence of randomness, the absence of a prolonged washout phase, the smaller number of hyperparameters, and the potential gain in speed compared to classical reservoir computing. In addition to these operational advantages, it required significantly less amount of training data in several examples \cite{barbosa2022learning}. 

Instead of using a static internal network in its core, the NG-RC uses a library of unique polynomials of time-shifted input variables of the input data $\textbf{u}(t)$ to generate the state vectors $\textbf{r}(t)$. 
As for classical reservoir computing only the linear output layer is optimized, leading to a highly efficient algorithm when deployed with a small state space dimension.

To learn the dynamics of a system, its input data $\mathbf{u}(t)$ is transformed with a polynomial multiplication library $\mathbf{P}^{O}$ and a time shift expansion $\mathbf{L}_k^s$ into a higher dimensional state space. Only the unique polynomials of specific orders $O$ are contained in $\mathbf{P}^{O}$ and are denoted by an index. To exemplify this notation, a two-dimensional input data point 
$\mathbf{u}(\tau) = (x(\tau), y(\tau))^T$ is considered and transformed with the unique polynomials of order 1 and 2 accordingly
\begin{equation}
    \mathbf{P}^{[1,2]}(\mathbf{u(\tau)}) = 
    \begin{pmatrix}
    x(\tau)\\[\jot]y(\tau)\\[\jot]x(\tau)^2\\[\jot]y(\tau)^2\\[\jot]x(\tau)y(\tau)
    \end{pmatrix}\
	\label{eq:quadratic}
\end{equation}
In addition, a time shift expansion $\mathbf{L}_k^s$ on the input data is applied. The data point to be expanded is concatenated with $k$ other past data points. The number of time steps between each of these is specified by the value $s$.
Including the time shift expansion $\mathbf{L}_{k=2}^{s=1}$ into this example generates the NG-RC state vector $\mathbf{r}(\tau+\Delta t)$
\begin{align}
    \mathbf{r}(\tau+\Delta t) &= \mathbf{P}^{[1,2]}(\mathbf{L}^{1}_{2}(\mathbf{v}(\tau)) \\
    &= \mathbf{P}^{[1,2]}(
    \begin{pmatrix}
    x(\tau)\\[\jot]y(\tau)\\[\jot]x(\tau-\Delta t)\\[\jot]y(\tau-\Delta t)
    \end{pmatrix}) =
    \begin{pmatrix}
    x(\tau)\\[\jot]y(\tau)\\[\jot]x(\tau-1)\\\vdots\\[\jot]x(\tau)y(\tau-\Delta t)\\[\jot]y(\tau)y(\tau-\Delta t)\\
    \end{pmatrix} \nonumber
    \\ \nonumber 
	\label{eq:quadratic}
\end{align}
Output is created by applying the readout matrix $\mathbf{W}_{out}$ to the state vector. Similarly to classical RC, $\mathbf{W}_{out}$ is the only component that needs to be optimized during the training process.
\\
\\
In the scope of this work, the NG-RC is optimized to become a one-step integrator, which generates the trajectory according to
\begin{eqnarray}
\textbf{v}(t+\Delta t) = \textbf{v}(t) + \textbf{W}_{out}\mathbf{r}(t + \Delta t)
\label{eq:output_ng}
\end{eqnarray}
 In the training phase, the input training data $\mathbf{X} = \mathbf{u}(0,\hdots,t_T)$ of length $T$ is transformed into the state matrix
 \begin{equation}
     \mathbf{R} = \mathbf{P}^{O}(\mathbf{L}^{s}_{k}(\mathbf{X})).
 \end{equation}
Consider that due to the time shift resulting from the choice of $k$ and $s$, a warm-up time $\delta t = ks$ is involved, where the entries of the state matrix are not defined for times $t < \delta t$. 
The output-target matrix $\mathbf{\tilde{X}}$ must be adjusted accordingly and is defined in terms of equation \ref{eq:output_ng} as follows
\begin{equation}
\mathbf{\tilde{X}} = \sum_{i=\delta t + 1}^{T} \mathbf{X}_i - \mathbf{X}_{i-1} = (\Delta \mathbf{v}_{\delta t+1}, \,\ldots \,, \Delta \mathbf{v}_T)^T
\label{eq:DifferenceNG_target}
\end{equation}
The readout matrix $\mathbf{W}_{out}$ is trained, equally as in classical reservoir computing, using ridge regression by minimizing
 \begin{eqnarray}
\sum^{}_{0 \leq t \leq T - \delta t} {\parallel  \textbf{W}_{out}\mathbf{R}_t - \mathbf{\tilde{X}}_t \parallel}^2 - \beta {\parallel \textbf{W}_{out} \parallel}^2 \ .
\label{eq:minimizing}
\end{eqnarray} 
In this setup, the NG-RC is trained to become a dynamical system that evolves according to 
\begin{equation}
    \mathbf{v}(t+\Delta t) = \mathbf{v}(t) + \mathbf{W}_{out} \mathbf{P}^{O}(\mathbf{L}^s_k(\mathbf{v}(t))).\\
\end{equation}

\subsection{Statistical measures}\label{measures}
When applying the control mechanism to nonlinear chaotic systems such as the Lorenz system, it does not make much sense to evaluate the effectiveness in terms of deviation from some concrete trajectory due to the very definition of chaos and its sensitive dependence on initial conditions. Rather, we would like to characterize the statistical long term properties of the dynamics, which is commonly referred to as the statistical "climate" of the system. It can be described by the following two quantitative measures:

\subsubsection{Correlation Dimension}
The correlation dimension is a fractal dimension measure, which characterizes the dimensionality of the space populated by the trajectory \cite{grassberger2004measuring}. It is related to the correlation integral 
\begin{eqnarray}
\begin{aligned}
C(r) &= \lim\limits_{N \rightarrow \infty}{\frac{1}{N^2}\sum^{N}_{i,j=1}\theta(r- | \textbf{x}_{i} - \textbf{x}_{j}  |)}      \\
       &= \int_{0}^{r} d^3 r^{\prime} c(\textbf{r}^{\prime}) \ ,
\label{eq:corrintegral}
\end{aligned}
\end{eqnarray}
where $\theta$ is the Heaviside function and $c(\textbf{r}^{\prime})$ is the standard correlation function. The correlation integral quantifies the mean probability that two states in phase space are close to each other at different time steps. The states are considered to be close to each other if their distance is less than the threshold distance $r$. The correlation dimension $\nu$ is then defined by the power-law relationship
\begin{eqnarray}
C(r) \propto r^{\nu} \ .
\label{eq:corrdim}
\end{eqnarray} 
For self-similar strange attractors, this relationship holds for a certain range of $r$, which therefore needs to be properly calibrated. As results in this study focus on a relative comparison, precision with regard to absolute values is not essential in this context. We use the Grassberger Procaccia algorithm \cite{grassberger1983generalized} to calculate the correlation dimension. \\

\subsubsection{Lyapunov Exponent}
Another aspect of the statistical climate is the temporal complexity of the attractor. It can be measured by its Lyapunov exponents $\lambda_{i}$, which describe the average rate of divergence of nearby points in phase space and thus measure sensitivity to initial conditions for each dimension. The magnitudes of $\lambda_{i}$ quantify the time scale on which the system becomes unpredictable \cite{wolf1985determining, shaw1981strange}. If the system exhibits at least one positive Lyapunov exponent, it is said to be chaotic. For our analysis, it is therefore sufficient to calculate only the largest Lyapunov exponent $\lambda_{max}$
\begin{eqnarray}
d(t) =  C e^{\lambda_{max} t} \ ,
\label{eq:lyapunov}
\end{eqnarray} 
which is calculated using the Rosenstein algorithm \cite{rosenstein1993practical}. It tracks the distance $d(t)$ of two initially nearby states in phase space, while the constant $C$ normalizes the initial separation. As for the correlation dimension, a relative comparison among different results is the goal and thus absolute precision is not of the highest importance. 

\section{Results}
\begin{figure} 
  \begin{center}
    \includegraphics[width=1.0\linewidth]{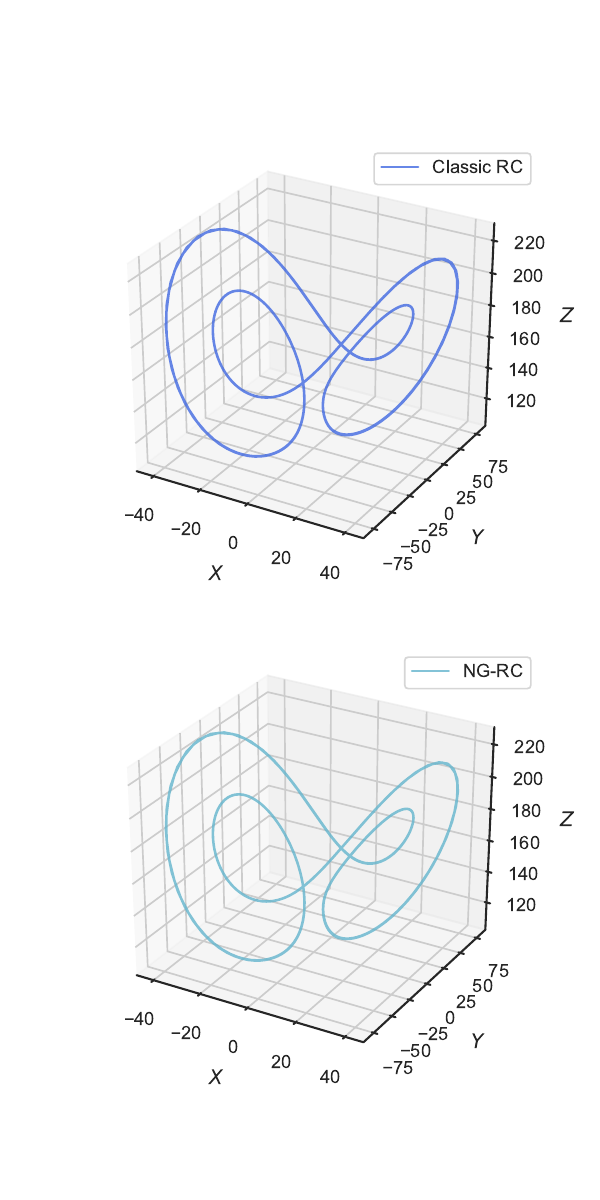}
    \caption{Top: Training data used for classic RC. Bottom: Training Data used for NG-RC.} 
    \label{fig:traindata}
  \end{center}
\end{figure}
As an example for a challenging control problem we investigate a situation where two complex dynamical states are involved. For this we initially simulate the Lorenz system such that it exhibits intermittent dynamics by setting parameters to $\bm{\pi}=[\sigma=10.0,\rho=166.15,\beta=8/3]$. Intermittent behavior is characterized by mostly periodic behavior, which is interrupted by irregular bursts occurring from time to time. Using the terminology from Section~\ref{sec:controlIntro}, we call this initial state $\textbf{X}$ and it is shown in the left plots of Fig~\ref{fig:chaos2intermittent}. As a next step, by changing $\rho$ to $\rho=167.2$ a new parameter set $\bm{\pi^{*}}$ is active that results in a chaotic state $\textbf{Y}$, which is represented by the middle plots. 

To apply the control mechanism, both classical RC and NG-RC are trained now using data of the original intermittent state $\textbf{X}$. In both cases, $N=5000$ training steps are used, while we allow for $t_{w}=1000$ washout steps for classical reservoir computing and a warm-up time of $\delta t = 57$ for NG-RC. The parameters used for classical RC are $\omega=0.0084$, $\rho=0.0084$ and $\beta=10^{-11}$. The NG-RC setup is $k=1$, $s=57$, $\beta=10^{-4}$ and $O=[1,2,3,4]$. While setting $k=1$ makes the NG-RC in principle equivalent to the SINDy algorithm \cite{brunton2016discovering}, except for the training routine, this choice turned out to work better than higher values for $k$. Because of the different warm-up/washout times, as well as different random seeds used, the simulated results of the Lorenz system in the intermittent state, as shown in the plots on the left, differ for classical RC and NG-RC. 
\begin{figure*}[t!] 
  \begin{center}
    \includegraphics[width=1.0\linewidth]{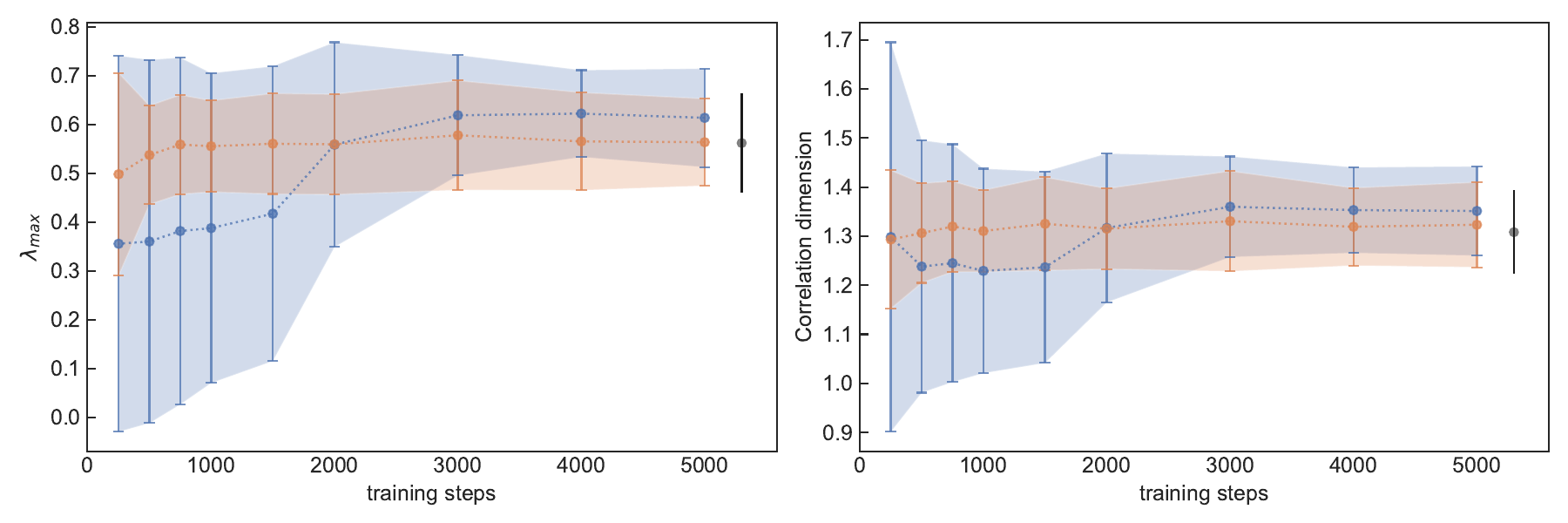}
    \caption{Left: Mean and standard deviation of the largest Lyapunov exponent of the controlled system for different amounts of training data based on 100 random realizations for classical RC (blue) and NG-RC (red). The grey reference bar represents the corresponding values for the simulated system in its original state $\textbf{X}$. Right: Similar representation for the correlation dimension. The x-axis values correspond to [250,500,750,1000,1500,2000,3000,4000,5000] training steps.} 
    \label{fig:trainerror}
  \end{center}
\end{figure*}

As a next step the control force as introduced in Section~\ref{sec:controlIntro} is applied based on the reservoir computing predictions, which tell how the system would have evolved if it was still in its original state $\textbf{X}$. The resulting attractors and trajectories are shown in the right plots of Fig~\ref{fig:chaos2intermittent}. In both cases (classical RC in the top two plots and NG-RC in the bottom two plots), the control is successful. The system is now forced back into an intermittent behavior, although the Lorenz system is still simulated using the parameter set $\bm{\pi^{*}}$, which leads to chaotic behavior in the absence of a control force. This is also confirmed when measuring the statistical climate of the attractors. For the classical RC variant, the original intermittent state $\textbf{X}$ has a largest Lyapunov exponent of $\lambda_{max}=0.624$ and correlation dimension $\nu=1.408$. In contrast, the chaotic state $\textbf{Y}$ has $\lambda_{max}=0.895$ and $\nu=1.744$. After applying the control force, the resulting values are $\lambda_{max}=0.477$ and $\nu=1.226$. For next-generation RC, the original state has $\lambda_{max}=0.599$ and $\nu=1.311$, while the chaotic state $\textbf{Y}$ has $\lambda_{max}=1.077$ and $\nu=1.628$ and the controlled state has $\lambda_{max}=0.570$ and $\nu=1.350$. Therefore the climate of the controlled state is very close to the original intermittent state. 

As next-generation RC is said to require less training data than classical RC, we investigate the same example for a significantly lower number of training steps $N=500$. The setup in terms of parameters is for both RC approaches the same as before, except that the washout phase of classical RC is reduced to $t_{w}=100$ steps in order to make the overall data need of the two methods more comparable. In general, even for classical RC already a relatively short washout phase of $t_{w}<<100$ is sufficient. The corresponding results are shown in Fig~\ref{fig:chaos2intermittent500}, which has the same structure as Fig~\ref{fig:chaos2intermittent}. However, while the control is still successful for NG-RC -- the lower right plots show the desired intermittent behavior as the original state in the lower left plots -- the results for classical RC look different. Here, the resulting attractor shows periodic dynamics and is missing the irregular bursts, which are characteristic for intermittent dynamics. An obvious assumption would be that this could happen when the relatively short training data does not include any of these bursts. Interestingly, this is the case for both variants, as Fig~\ref{fig:traindata} suggests. However, NG-RC still captures that the system is actually in an intermittent state, even in the absence of bursts. Understanding why this is the case for NG-RC but not for classical RC will be the subject of future research.

In addition, a statistical analysis is carried out by repeating the same procedure for $100$ different random seeds that lead to different random realizations of the reservoir $\textbf{A}$ and $\textbf{W}_{in}$, as well as different starting points and thus trajectories for the intermittent Lorenz system. Figure~\ref{fig:trainerror} shows the results in terms of the mean and standard deviation of the largest Lyapunov exponent and the correlation dimension for different amounts of training data steps. The grey bar represents the mean and standard deviation of the results evaluated for the simulated Lorenz system being in its initial intermittent state $\textbf{X}$. It is clearly visible that NG-RC (red) outperforms classical RC (blue), particularly for 2000 and less training data steps. For 500 training steps, NG-RC still delivers very good results and can control the Lorenz system into its intermittent target state with high precision in terms of statistical climate, while classical RC fails to do so reliably. This statistically confirms the results shown as an example in Fig~\ref{fig:chaos2intermittent500}. If the training data is even further shortened to 250 data points, the standard deviation for NG-RC becomes slightly larger, and the mean values decline. However, the statistical climate is still reasonably close to the desired target state. The results are equivalent when looking at the same measures evaluated on the respective predictions of both approaches (not shown here). As a reference, the measures for the chaotic state $\textbf{Y}$, which the control method is supposed to avoid, are $0.845$ and $0.066$ for mean and standard deviation of $\lambda_{max}$ as well as $1.690$ and $0.065$ for the correlation dimension, respectively. Heuristically one can state that using RC-NG instead of classical reservoir computing in this context requires roughly 10 times less data.

\section{Outlook}
The model-free and data-driven methods presented here can generally be used to control dynamical systems to arbitrary states for which only limited data is available. The NG-RC excels at this task and requires significantly less training data than classical RC. This adds another viable control method and expands the scope of application. A worthwhile direction for further research would be whether a non-stationary process can be controlled into an arbitrarily corresponding stationary process. The less training data is needed, the more it can be approximated as a stationary training sample.

\section*{Acknowledgment}
We would like to thank the DLR and Allianz Global Investors for providing data and computational resources.

\bibliography{bibliography}

\providecommand{\noopsort}[1]{}\providecommand{\singleletter}[1]{#1}%
\begin{thebibliography}{10}

\bibitem{barbosa2022learning}
Wendson~AS Barbosa and Daniel~J Gauthier.
\newblock Learning spatiotemporal chaos using next-generation reservoir
  computing.
\newblock {\em Chaos: An Interdisciplinary Journal of Nonlinear Science},
  32(9):093137, 2022.

\bibitem{boccaletti2000control}
Stefanos Boccaletti, Celso Grebogi, Y-C Lai, Hector Mancini, and Diego Maza.
\newblock The control of chaos: theory and applications.
\newblock {\em Physics reports}, 329(3):103--197, 2000.

\bibitem{brunton2016discovering}
Steven~L Brunton, Joshua~L Proctor, and J~Nathan Kutz.
\newblock Discovering governing equations from data by sparse identification of
  nonlinear dynamical systems.
\newblock {\em Proceedings of the national academy of sciences},
  113(15):3932--3937, 2016.

\bibitem{carroll2022time}
Thomas~L Carroll and Joseph~D Hart.
\newblock Time shifts to reduce the size of reservoir computers.
\newblock {\em Chaos: An Interdisciplinary Journal of Nonlinear Science},
  32(8):083122, 2022.

\bibitem{erdos1959random}
Paul Erdos.
\newblock On random graphs.
\newblock {\em Publicationes mathematicae}, 6:290--297, 1959.

\bibitem{gallicchio2018design}
Claudio Gallicchio, Alessio Micheli, and Luca Pedrelli.
\newblock Design of deep echo state networks.
\newblock {\em Neural Networks}, 108:33--47, 2018.

\bibitem{gauthier2021next}
Daniel~J Gauthier, Erik Bollt, Aaron Griffith, and Wendson~AS Barbosa.
\newblock Next generation reservoir computing.
\newblock {\em Nature communications}, 12(1):5564, 2021.

\bibitem{grassberger1983generalized}
Peter Grassberger.
\newblock Generalized dimensions of strange attractors.
\newblock {\em Physics Letters A}, 97(6):227--230, 1983.

\bibitem{grassberger2004measuring}
Peter Grassberger and Itamar Procaccia.
\newblock Measuring the strangeness of strange attractors.
\newblock In {\em The Theory of Chaotic Attractors}, pages 170--189. Springer,
  2004.

\bibitem{haluszczynski2020reducing}
Alexander Haluszczynski, Jonas Aumeier, Joschka Herteux, and Christoph
  R{\"a}th.
\newblock Reducing network size and improving prediction stability of reservoir
  computing.
\newblock {\em Chaos: An Interdisciplinary Journal of Nonlinear Science},
  30(6):063136, 2020.

\bibitem{haluszczynski2021controlling}
Alexander Haluszczynski and Christoph R{\"a}th.
\newblock Controlling nonlinear dynamical systems into arbitrary states using
  machine learning.
\newblock {\em Scientific reports}, 11(1):12991, 2021.

\bibitem{herteux2020reservoir}
Joschka Herteux and Christoph R{\"a}th.
\newblock Reservoir computing and its sensitivity to symmetry in the activation
  function.
\newblock {\em arXiv preprint arXiv:2010.07103}, 2020.

\bibitem{jaeger2001echo}
Herbert Jaeger.
\newblock The ``echo state'' approach to analysing and training recurrent
  neural networks-with an erratum note.
\newblock {\em Bonn, Germany: German National Research Center for Information
  Technology GMD Technical Report}, 148(34):13, 2001.

\bibitem{jaeger2004harnessing}
Herbert Jaeger and Harald Haas.
\newblock Harnessing nonlinearity: Predicting chaotic systems and saving energy
  in wireless communication.
\newblock {\em science}, 304(5667):78--80, 2004.

\bibitem{kong2021machine}
Ling-Wei Kong, Hua-Wei Fan, Celso Grebogi, and Ying-Cheng Lai.
\newblock Machine learning prediction of critical transition and system
  collapse.
\newblock {\em Physical Review Research}, 3(1):013090, 2021.

\bibitem{lorenz1963deterministic}
Edward~N Lorenz.
\newblock Deterministic nonperiodic flow.
\newblock {\em Journal of atmospheric sciences}, 20(2):130--141, 1963.

\bibitem{maass02}
Wolfgang Maass, Thomas Natschlaeger, and Henry Markram.
\newblock Real-time computing without stable states: A new framework for neural
  computation based on perturbations.
\newblock {\em Neural Computation}, 14(11):2531--2560, 2002.

\bibitem{maass2002real}
Wolfgang Maass, Thomas Natschl{\"a}ger, and Henry Markram.
\newblock Real-time computing without stable states: A new framework for neural
  computation based on perturbations.
\newblock {\em Neural computation}, 14(11):2531--2560, 2002.

\bibitem{ott1990controlling}
Edward Ott, Celso Grebogi, and James~A Yorke.
\newblock Controlling chaos.
\newblock {\em Physical review letters}, 64(11):1196, 1990.

\bibitem{pyragas1992continuous}
Kestutis Pyragas.
\newblock Continuous control of chaos by self-controlling feedback.
\newblock {\em Physics letters A}, 170(6):421--428, 1992.

\bibitem{rosenstein1993practical}
Michael~T Rosenstein, James~J Collins, and Carlo~J De~Luca.
\newblock A practical method for calculating largest lyapunov exponents from
  small data sets.
\newblock {\em Physica D: Nonlinear Phenomena}, 65(1-2):117--134, 1993.

\bibitem{schiff1994controlling}
Steven~J Schiff, Kristin Jerger, Duc~H Duong, Taeun Chang, Mark~L Spano, and
  William~L Ditto.
\newblock Controlling chaos in the brain.
\newblock {\em Nature}, 370(6491):615--620, 1994.

\bibitem{shaw1981strange}
Robert Shaw.
\newblock Strange attractors, chaotic behavior, and information flow.
\newblock {\em Zeitschrift f{\"u}r Naturforschung A}, 36(1):80--112, 1981.

\bibitem{shinbrot1993using}
Troy Shinbrot, Celso Grebogi, James~A Yorke, and Edward Ott.
\newblock Using small perturbations to control chaos.
\newblock {\em nature}, 363(6428):411--417, 1993.

\bibitem{wolf1985determining}
Alan Wolf, Jack~B Swift, Harry~L Swinney, and John~A Vastano.
\newblock Determining lyapunov exponents from a time series.
\newblock {\em Physica D: Nonlinear Phenomena}, 16(3):285--317, 1985.

\end{thebibliography}
\bibliographystyle{plain}

\end{document}